\documentclass[12pt,journal,onecolumn]{IEEEtran}

\usepackage{amsmath,amssymb,amsfonts,amsthm}
\usepackage{cite}
\usepackage{url}
\usepackage{graphicx}

\newcommand\normx[1]{\left\Vert#1\right\Vert}

\begin{document}

\title{Loss Regularizing Robotic Terrain Classification}

\author{
Shakti~Deo~Kumar,~Sudhanshu~Tripathi,~Krishna~Ujjwal,~Sarvada~Sakshi~Jha, and~Suddhasil~De~(\IEEEmembership{Senior~Member,~IEEE})\\%
Department of Computer Science and Engineering,\\%
National Institute of Technology Patna,~India\\%
Email: suddhasil.de@acm.org\\%
{\footnotesize{Note: Preliminary draft of the work published in IEEE conference 2023.}}
}

\maketitle

\begin{abstract}
Locomotion mechanics of legged robots are suitable when pacing through difficult terrains. Recognising terrains for such robots are important to fully yoke the versatility of their movements. Consequently, robotic terrain classification becomes significant to classify terrains in real time with high accuracy. The conventional classifiers suffer from overfitting problem, low accuracy problem, high variance problem, and not suitable for live dataset. On the other hand, classifying a growing dataset is difficult for convolution based terrain classification. Supervised recurrent models are also not practical for this classification. Further, the existing recurrent architectures are still evolving to improve accuracy of terrain classification based on live variable-length sensory data collected from legged robots. This paper proposes a new semi-supervised method for terrain classification of legged robots, avoiding preprocessing of long variable-length dataset. The proposed method has a stacked Long Short-Term Memory architecture, including a new loss regularization. The proposed method solves the existing problems and improves accuracy. Comparison with the existing architectures show the improvements.
\end{abstract}

\section{Introduction}\label{sec:Intro}

Legged robots are mobile in nature by virtue of locomotion using bio-inspired mechanics of their articulated movable limbs. They are useful for operations in difficult terrains and other challenging conditions, due to their adaptable and flexible locomotion strategies. Due to their instantaneous adapting ability to slope, roughness, undulation and other terrain obstacles, legged robots provide promising benefits when pacing through difficult terrains in comparison to wheeled robots. This is accomplished by altering their gaits, foot-tip arc shapes, footfall positions, stride length, etc. to adjust their behaviour and manage the difficulties posed by their surrounding terrains.

Terrain classification is an important step in fully yoking the versatility in movements of autonomous legged robots to explore difficult terrains. Knowledge about terrain provides individual robots with ability to correctly identify terrain types, and so has the potential to adapt to harness their versatile movements and overcome the difficulty faced when walking and in maintaining balance on such terrains. Consequently, the robotic terrain classification technique becomes significant to characterize and classify terrains in real time with high accuracy.

Since robotic terrain classification is a classification problem, machine learning based classification approaches are used to solve the problem. The conventional classifiers, like decision tree, logistic regression, support vector machine (SVM)~\cite{enwiki:1190739318}, etc., as well as the neural network classifiers pose diverse shortcomings while solving this classification problem. The main problem with conventional classifier is that there is only one unit that is producing the activation output. It will try to fit a linear model for classification. Any attempt to use quadratic or polynomial components for the line might improve training accuracy but also results in overfitting most of the time. As a result, the accuracy metric can reach only a tangential or asymptotic maximum to around 70\% to 85\%. In other words, high variances are possible in such solutions. Further, the length of data must be predetermined in order to extract time-domain characteristics by these models, and so they are difficult to handle datasets with variable time durations.

On the other hand, in a neural network there can be multiple units in a single layer, and all of those units individually are classifiers with some activation function. Additionally, there can be multiple such layers. A major advantage of having such a collection of classifiers is that each of those classifiers can engage themselves for finding different patterns in the data. Since there are many such units, the network can learn a lot of features from the data. Artificial neural network (ANN) classifier has a group of multiple neurons at each layer, with a collection of three types of layers – input layer, hidden layer and output layer. Respective functions of these layers include accepting the inputs, learning from the inputs, and producing the classification result based on inputs. Essentially, each of these layer contribute towards updating trainable parameters. Convolutional neural network (CNN) classifier~\cite{enwiki:1197047138,dhanraj2019efficient} is based on the computational model that uses a variation of multilayer neurons~\cite{enwiki:1198631061}. It contains one or more convolutional layers, either entirely connected or pooled, for creating feature maps, which are used to record certain portions of input data, commonly, imagery data. The feature maps in turn are split rectangularly, and are then transferred for processing in nonlinear fashion to detect important features and patterns without any human intervention. However, live training and classifying with growing dataset is difficult for CNN based robotic terrain classification.

Recurrent Neural Network (RNN)~\cite{enwiki:1214097285} has been employed through supervised learning mechanism for robotic terrain classification, due to its ability to extract temporal properties from captured time-series sensory data. Such models immediately employ raw sensory stream as input instead of gathering across time and transferring to suitable descriptors, and as a result they are considered more appropriate for real-time classification. However, gathering data for supervised models is a laborious task. Either the robots must traverse various terrains independently for data annotation, or the data must be hand-labeled later. Consequently, semi-supervised approaches~\cite{kumar2023robotic} for terrain classification are prominent, since the effort may be decreased if just a piece of the data has to be annotated for equivalent performance. However, the existing RNN architectures are still evolving to improve accuracy of terrain classification based on live variable-length sensory data collected from legged robots.

In this paper, a new semi-supervised method is proposed for terrain classification based on live variable-length data from force sensors and inertial measurement unit (IMU) sensors affixed to legged robots. Working with live-streamed sensory data makes sure that the tedious and challenging task of preprocessing the long dataset of variable length is avoided. The proposed method has a stacked architecture of Long Short-Term Memory (LSTM) model~\cite{enwiki:1213969171}, in which hierarchical arrangement of unsupervised and supervised learning mechanisms improve accuracy while using smaller quantity of annotated data than the supervised approach. Further, a new loss regularization approach is also included with the proposed method to correctly recognise terrain on which legged robot is moving. Comparison with the other existing classifier architectures show the improvements of the proposed method.

The rest of the paper is arranged as follows. Section~\ref{sec:Proposed} introduces the proposed method, and elaborates its detailed functioning along with mathematical formulation. Section~\ref{sec:exp_res} describes the dataset, and presents the experiment details and the results obtained from the experimentation. A discussion about the comparison with the existing techniques is also added in the section. Section~\ref{sec:concl} concludes the paper.

\section{Proposed method}\label{sec:Proposed}

This section presents the proposed semi-supervised method to recognise terrain on which legged robot is moving based on its force sensor and inertial measurement unit (IMU) sensor recorded historical motion data. The objective of the proposed method is to provide an estimate of robot's force and motion actuator data in step at time $t$, which can be made from prediction of terrain by observing force and motion data of its previous steps up to time $t-1$. The method further considers limited amount of labelled data for its learning. The proposed method is built as stacked architecture of a deep layered organization of LSTM model.

The proposed semi-supervised method comprises of two LSTM layers, that are stacked on top of one another. The first LSTM layer of the proposed method consists of 200 hidden units, and it receives the input data. The responsibility of the first LSTM layer is to act as predicting layer based on unsupervised learning technique on a small portion of dataset, where it inputs $X_{1:t-1}$ and $X_t$ (for all $t=1~\text{to}~T$) for training and validation. Post-training, the weights of the first LSTM layer are preserved and not allowed to change. The layer estimates the data ${\overset{~_p}{\widehat{X}}}_t$ as output using the previous data $X_{t-1}$ as input. The second LSTM layer, which also consists of 200 hidden units, is layered above the first LSTM layer and is provided with the remaining part of dataset for training. While training with validation, the second LSTM layer inputs data $X_{t}$ and $Y_{t}$ (for all $t=1~\text{to}~T$) from dataset and its weights are thereby adjusted as per the classifying objectives of terrain recognition. Post-training, this layer takes the data ${\overset{~_p}{\widehat{X}}}_t$ and estimates the classification accordingly as output ${\overset{\,_c}{\widehat{Y}}}_t$. The second LSTM layer is responsible for classifying data based on predicting capability, as output from the first LSTM layer goes to the second LSTM layer. The arrangement of LSTM layers has the benefit of better ability to handle the temporal dependency of input time series data. Further, enough memory cells provide storing ability of large data sequence. Both the LSTM layers do not enforce the need for windowing data. The proposed method ends with a fully connected (FC) layer, having one hidden unit with linear activation, and corresponds to the final expected output of thrust or inertial data for the following time step. Additional benefit of the proposed method is its adaptability, in which training of the method is initiated with small portion of data, and with availability of more data, the dataset is enriched to gradually improve performance.

The predicting loss function ${\overset{~_p}{\widehat{\mathcal{L}}}}$ is defined to be the mean squared error between \(X_t\) and \({\overset{~_p}{\widehat{X}}}_t\) where \(X_t\) represents the processed sensor data directly from dataset at time $t$ and \({\overset{~_p}{\widehat{X}}}_t\) is the predicted estimate by the first LSTM layer about \(X_t\), where, $X_{t_j}$ is the processed sensor data directly from dataset to give as input to the $j$-th hidden unit at $t$, ${\overset{~_p}{\widehat{X}}}_{t_j}$ is the predicted estimate as output from the $j$-th hidden unit at $t$, and $h$ is the hidden unit count.

\begin{align}
   {\overset{~_p}{\widehat{\mathcal{L}}}}=&\frac{1}{h}\sum_{j=1}^{h}\Big(X_{t_j}-{\overset{~_p}{\widehat{X}}}_{t_j}\Big)^{2}+\lambda\bigg(\gamma\!\!\!\!\sum_{\substack{\rho\in\\\{\mathsf{W},\mathsf{U},\mathsf{B}\}}}\!\!\!\normx{{\overset{~_p}{\hat{\theta}}}_{\!\!\rho}}_{1}+(1-\gamma)\!\!\!\sum_{\substack{\rho\in\\\{\mathsf{W},\mathsf{U},\mathsf{B}\}}}\!\!\!\normx{{\overset{~_p}{\hat{\theta}}}_{\!\!\rho}}^{2}_{2}\bigg)\label{eqn:lstmpredictingloss}
\end{align}
\noindent The hyperparameter $\lambda$ (where, $\lambda\in\mathbb{R}$) is a regularization parameter that can be changed. By altering the value of $\lambda$, the neutralising effect between the loss function term and regularizing term is kept in balance. $\|{{\overset{~_p}{\hat{\theta}}}}\|_{1}$ and $\|{{\overset{~_p}{\hat{\theta}}}}\|^{2}_{2}$ respectively denote the $\ell^1$-norm and $\ell^2$-norm for the trainable parameters $\mathsf{W}$, $\mathsf{U}$ and $\mathsf{B}$ of the first LSTM layer. The hyperparameter $\gamma$ (where, $\gamma\in[0,1]$) denotes the proportionality of $\ell^1$-norm and $\ell^2$-norm in the regularization.
\begin{figure}[t]
\centering
\includegraphics[width=0.55\linewidth]{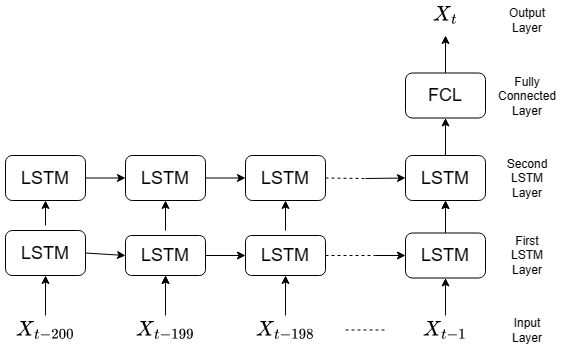}
\caption{Proposed semi-supervised method's architecture.}\label{fig:proparch}
\end{figure}

The classifying loss function ${\overset{~_c}{\widehat{\mathcal{L}}}}$ is defined to be the cross entropy in relation to \(Y_t\) and \({\overset{\,_c}{\widehat{Y}}}_t\), where \(Y_t\) represents the terrain class data directly from dataset at time $t$ and \({\overset{\,_c}{\widehat{Y}}}_t\) is the classified estimate by the second LSTM layer about \(Y_t\).
\begin{align}
   {\overset{~_c}{\widehat{\mathcal{L}}}}=&\sum_{i=1}^{\mathfrak{C}}\Big(Y_i\log\big({\overset{\,_c}{\widehat{Y}}}_i\big)\Big)+\lambda\bigg(\gamma\!\!\!\!\sum_{\substack{\rho\in\\\{\mathsf{W},\mathsf{U},\mathsf{B}\}}}\!\!\!\normx{{\overset{~_c}{\hat{\theta}}}_{\!\!\rho}}_{1}+(1-\gamma)\!\!\!\sum_{\substack{\rho\in\\\{\mathsf{W},\mathsf{U},\mathsf{B}\}}}\!\!\!\normx{{\overset{~_c}{\hat{\theta}}}_{\!\!\rho}}^{2}_{2}\bigg)\label{eqn:lstmclassifyingloss}
\end{align}
where, \(\mathfrak{C}\) is the different types of terrains considered. For Eq.~\eqref{eqn:lstmclassifyingloss}, every dimension of data from dataset is normalised during pre-processing to result in mean value zero and standard deviation of one. $\|{{\overset{~_c}{\hat{\theta}}}}\|_{1}$ and $\|{{\overset{~_c}{\hat{\theta}}}}\|^{2}_{2}$ respectively denote the $\ell^1$-norm and $\ell^2$-norm for the trainable parameters $\mathsf{W}$, $\mathsf{U}$ and $\mathsf{B}$ of the second LSTM layer. The hyperparameters $\lambda$ and $\gamma$ have the responsibilities similar to Eq.~\eqref{eqn:lstmpredictingloss}.

The adaptive moment estimation (Adam) optimizer is used for optimizing learning rate in adaptive fashion. Adam makes use of an exponentially moving average to gauge the moments of the gradient, viz. mean of gradient and element-wise square of gradient. The proposed method uses Adam due to low count of hyperparameters. For each weight parameter of the proposed method, individual learning rate is maintained and adjusted independently with training. Rectified linear unit (ReLU) function is used as activation function in the input layer of the proposed method. Softmax function is used in output layer of the proposed method for exponential activation to generate the probabilistic distribution of outcome corresponding to all terrain types. The outcome with maximum probability in the distribution is considered as the final output.

\section{Experiments \& Results}\label{sec:exp_res}

The proposed method described in Section~\ref{sec:Proposed} has the objective of classifying terrain that the robot is walking on based on time series dataset, with history of data gathered from sensors attached to the robot. The experimentations of the proposed method is performed on the benchmark time series dataset for training and testing.

\subsection{Dataset}

\begin{table}[tb]
\caption{Parameter list for the proposed method.}\label{tab:expsettings}
\begin{center}
\begin{tabular}{@{}l@{~} @{~}l@{}}
\hline
$\qquad$Parameters & $\quad$Values$\quad$\\%
\hline
      $\quad$Epochs & $\quad$300\\
      $\quad$Dropout & $\quad$20\%\\
      $\quad$$k$ in $k$-fold validation & $\quad$5\\
      $\quad$Learning rate & $\quad$0.005\\
      $\quad$Batch size & $\quad$512\\
\hline
\end{tabular}
\end{center}
\end{table}
QCAT dataset is used to train and test the proposed method. The dataset is based on robotic walking sensor data patten over six exterior terrains, viz. concrete, grassy, gravel, mulch, dirt, sandy. The walking pattern involves six distinct speed per exterior terrain, while walking with uniform gait yet three varied footstep frequencies and two different pace lengths. Data are gathered for eighty footsteps per terrain and per pace. In each recording, data is collected from measurements of --- (i) four force sensors affixed to quadruped's feet (viz. forward left foot, forward right foot, back left foot, back right foot) covering twelve dimensions, and (ii) the IMU sensor covering three linear accelerations from accelerometer, three angular velocities from gyroscope, and four orientations for all directions. The collected robot's force and motion data in the dataset is spanned across multiple CSV structures, which are grouped based on terrain type and speed of robot on that terrain. Combining the CSV structures results in a multi-dimensional array with all the features of the force and motion data and the corresponding terrain. Normalization of the multi-dimensional array is carried out to generate standard deviation and mean values one and zero respectively. The entire dataset is randomly split into two parts: 90\% of the data are used by the first LSTM layer for predicting, while the remaining 10\% are used by the second LSTM layer and the fully connected layer for classifying. During testing, randomly 10\% of the data are selected from the preceding 90\% split.

\subsection{Training and testing}
Typically, the training of any learning model is thought of as optimisation of corresponding loss function. In this paper, Eq.~\eqref{eqn:lstmpredictingloss} is optimized during training of the first LSTM layer, while Eq.~\eqref{eqn:lstmclassifyingloss} is optimized while training of the second LSTM layer. The learning rate is taken as 0.005, which is decided upon experimentally after doing several experiments. After each LSTM layer in the hierarchical architecture, 20\% dropout is enforced to regularize and stop the proposed method from over-fitting during training. The batch size is taken as 512 training samples. Other experimental parameters are found analytically.

\begin{figure}[t]
\centering
\includegraphics[width=0.45\linewidth]{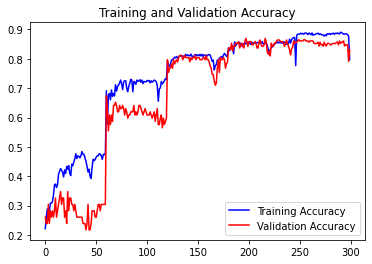}
\caption{Plot of training and validation accuracy against epochs.}\label{fig:propaccplot}
\end{figure}
\begin{figure}[t]
\centering
\includegraphics[width=0.45\linewidth]{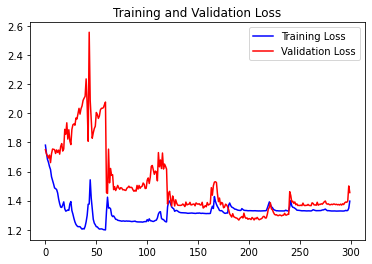}
\caption{Plot of training and validation losses against epochs.}\label{fig:proplossplot}
\end{figure}

During training, the first LSTM layer is trained and validated in 80\%-20\% split with 90\% of the data from QCAT dataset without any windowing. The second LSTM layer and the fully connected layer are trained by only 5\% of the QCAT data, and validated on the rest 5\% of the dataset. While training, $k$-fold validation is performed to reduce the bias effect to as low as possible. The benefit of $k$-fold validation is due to the independence among each other of the training data samples in QCAT dataset. Fig.~\ref{fig:propaccplot} shows the improvement in training and validation accuracy with the increase in epochs. The figure also shows the stable training and validation outcomes after epoch threshold, even though input data is highly varying in nature. Fig.~\ref{fig:proplossplot} shows the effect of loss regularization with the rise in epochs. In this figure also, it is observed that both the training and validation loss patterns are converging with more of epochs for the proposed method. Both Fig.~\ref{fig:propaccplot} and Fig.~\ref{fig:proplossplot} shows 200 epochs as the best performance of the proposed method with maximum training and validation accuracy and minimum training and validation losses respectively.

\subsection{Results}
Overall findings depict the effectiveness of the proposed method as learning model when few labeling is provided, notably for terrain identification, where substantial annotations of dataset may be challenging. Table~\ref{tab:expComp} compares the nature and the accuracy results of the proposed method. The table shows the accuracy of around 89\%, which produces significant improvement of 22\% over SVM based technique and 9\% over fully convolutional network (FCN) based technique. The proposed method also even improves by 1.5\% over temporal convolutional network (TCN) based technique.
\begin{table}[tb]
\caption{Comparison of the proposed method with the existing works.}\label{tab:expComp}
\begin{center}
\begin{tabular}{@{}l@{~} @{$\quad$}c@{~} @{~}c@{}}
\hline
$\quad$Work & Nature & Accuracy\\%
\hline
      ~SVM & Simple & 67\%\\
      ~FCN & Simple & 80.39\%\\
      ~TCN & Simple & 87.5\%\\
      ~Proposed method & Stacked & 89.04\%\\
\hline
\end{tabular}
\end{center}
\end{table}

\section{Conclusion}\label{sec:concl}

This paper has presented a new semi-supervised method for terrain classification of legged robots. The proposed method has contributed twofold, viz. by adopting a stacked LSTM architecture, and by including a new loss regularization approach. The method is experimented on variable-length QCAT dataset of time-series sensory data, which are collected for concrete, grassy, gravel, mulch, dirt and sandy exterior terrains. During experimentation, training and validation accuracy, as well as training and validation losses are found to be compatible with rise in epochs. The results have further shown sufficient improvements in accuracy of the proposed method over baseline models. These developments have made possible their use supplementing the state-of-the-art in emergency scenarios~\cite{pan21,pan24}.

\bibliographystyle{unsrt}
\bibliography{citation}

\end{document}